\documentclass[11pt]{article}
\usepackage{ijcai17}
\usepackage{times}
\usepackage{url}
\usepackage{latexsym}

\usepackage{graphicx}
\usepackage{amsmath}
\usepackage{multirow}
\usepackage{booktabs}
\usepackage{array}
\usepackage{subfigure}
\usepackage{amssymb}

\title{Image-embodied Knowledge Representation Learning}
\author{Ruobing Xie$^{1}$, Zhiyuan Liu$^{1,2}$\thanks{Corresponding author: Z. Liu (liuzy@tsinghua.edu.cn)}, Huanbo Luan$^{1}$, Maosong Sun$^{1,2}$\\
$^{1}$ Department of Computer Science and Technology, \\
State Key Lab on Intelligent Technology and Systems, \\
National Lab for Information Science and Technology, Tsinghua University, Beijing, China \\
$^{2}$ Jiangsu Collaborative Innovation Center for Language Ability, \\
Jiangsu Normal University, Xuzhou 221009 China \\
}

\date{}

\begin{document}
\maketitle
\begin{abstract}
  Entity images could provide significant visual information for knowledge representation learning. Most conventional methods learn knowledge representations merely from structured triples, ignoring rich visual information extracted from entity images. In this paper, we propose a novel Image-embodied Knowledge Representation Learning model (IKRL), where knowledge representations are learned with both triple facts and images. More specifically, we first construct representations for all images of an entity with a neural image encoder. These image representations are then integrated into an aggregated image-based representation via an attention-based method. We evaluate our IKRL models on knowledge graph completion and triple classification. Experimental results demonstrate that our models outperform all baselines on both tasks, which indicates the significance of visual information for knowledge representations and the capability of our models in learning knowledge representations with images.
\end{abstract}

\section{Introduction}

Knowledge graphs (KGs), which provide huge amount of structured information for entities and relations, have been successfully utilized in various fields such as knowledge inference \cite{yang2014embedding} and question answering \cite{yin2016neural}. A typical KG like Freebase or DBpedia usually models the multi-relational information with enormous triple facts represented as $(\emph{head entity}, \texttt{relation}, \emph{tail entity})$, which is also abridged as $(h,r,t)$.

Recently, translation-based methods are proposed to model knowledge graphs. These methods project both entities and relations into a continuous low-dimensional semantic space, with relations considered to be translating operations between head and tail entities \cite{bordes2013translating}. Translation-based methods could leverage both effectiveness and efficiency in knowledge representation learning (KRL), and thus have attracted great attention in recent years. However, most conventional methods on KRL only concentrate on the structured information in triple facts, regardless of rich external information located in entity images.

\begin{figure}[!htbp]
\centering
\includegraphics[width=0.9\columnwidth]{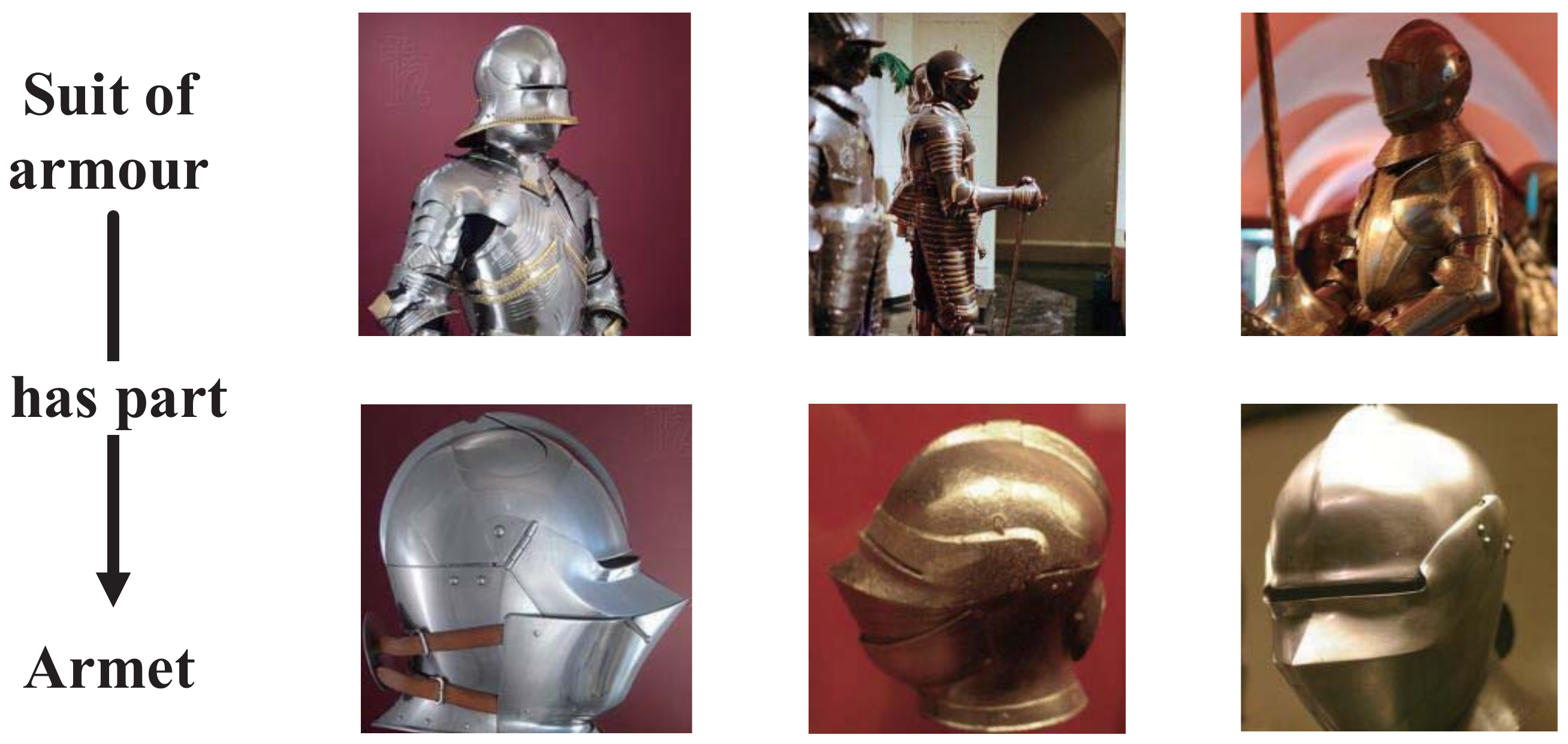}
\caption{Examples of entity images.}\label{fig. 1}
\end{figure}

Fig. \ref{fig. 1} demonstrates some examples of entity images. Each entity has multiple images which can provide significant visual information that intuitively describes the appearances and behaviours of this entity. To utilize the rich information in images, we propose the Image-embodied Knowledge Representation Learning model (IKRL). More specifically, we first propose an image encoder which consists of a neural representation module and a projection module to generate the image-based representation for each image instance. Second, we construct the aggregated image-based representation for each entity jointly considering all its image instances with an attention-based method. Finally, we jointly learn the knowledge representations with translation-based methods.

We evaluate the IKRL model on knowledge graph completion and triple classification. Experimental results demonstrate that our model achieves the state-of-the-art performances on both tasks, which confirms the significance of visual information in knowledge representation learning. It also indicates that our IKRL model is capable of encoding image information well into knowledge representations. We demonstrate the main contributions of this work as follows:
\begin{itemize}
  \item We propose a novel IKRL model considering visual information in entity images for knowledge representation learning. To the best of our knowledge, this is the first attempt to combine images with knowledge graphs for knowledge representation learning.
  \item We evaluate our models on a real-world dataset and receive promising performances on both knowledge graph completion and triple classification.
  \item We further conduct detailed analysis on representative cases, which confirms the power of attention in selecting informative images. We also find some interesting semantic regularities between image representations.
\end{itemize}

\section{Related Work}

\subsection{Translation-based Methods}

Translation-based methods have achieved great success on knowledge representation learning in recent years. TransE \cite{bordes2013translating} models both entities and relations into the same low-dimensional continuous vector space, with relations considered to be translating operations between head and tail entities. The basic assumption of TransE is that the embedding of tail entity $\mathbf{t}$ should be the neighbour of $\mathbf{h}+\mathbf{r}$. The energy function of TransE is defined as follows:
\begin{equation}
\begin{split}
E(h,r,t)=||\mathbf{h}+\mathbf{r}-\mathbf{t}||.
\end{split}
\end{equation}
TransE is both effective and efficient, while the simple assumption may result in conflicts when modeling complicated entities and relations. To address this problem, TransH \cite{wang2014transH} proposes relation-specific hyperplanes for translations between entities. TransR \cite{lin2015learning} models entities and relations in different vector spaces, projecting entities from entity space to relation spaces with relation-specific matrices. TransD \cite{ji2015knowledge} further considers the diversities of both entities and relations, using dynamic mapping matrix for the multiple representations of entities. However, these methods only concentrate on the structured information in KGs. We propose the IKRL model to consider images based on TransE, and our model can also be easily extended to other translation-based methods.

\subsection{Multi-source Information Learning}

Multi-source information such as textual and visual information is significant for knowledge representation. To utilize rich textual information, \cite{wang2014knowledge} projects both entities and words into a joint vector space with alignment models. \cite{xie2016representation} directly constructs entity representations from entity descriptions, which is capable of modeling new entities. As for visual information, multimodal representations based on words and images are widely used on various tasks like image-sentence ranking \cite{kiros2014unifying}, metaphor identification \cite{shutovablack2016black} and visual question answering \cite{antol2015vqa}. However, image information has not yet been used in knowledge representations. To the best of our knowledge, IKRL is the first attempt which explicitly encodes visual information from images into knowledge representations.

\section{Methodology}

We first introduce the notations used in this paper. Given a triple $(h,r,t) \in T$, it consists of two entities $h,t \in E$ and a relation $r \in R$. $T$ stands for the whole training set of triples, $E$ represents the set of entities, and $R$ represents the set of relations. Each entity and relation embedding takes value in $\mathbb{R}^{d_s}$ with $d_s$ to be the dimension.

To utilize entity image information in KRL, we propose two kinds of representations for each entity. We set $\mathbf{h}_S$, $\mathbf{t}_S$ as the \textbf{structure-based representations} (SBR) of head and tail entities, which are the distributed representations learned by conventional KRL models. We also propose a novel kind of knowledge representations $\mathbf{h}_I$, $\mathbf{t}_I$ as the \textbf{image-based representations} (IBR), which are constructed from the corresponding images of head and tail entities.

\subsection{Overall Architecture}

We attempt to utilize structured knowledge information as well as visual information in the IKRL model. Following the framework of translation-based methods, we define the overall energy function as follows:
\begin{equation}
\begin{split}
E(h,r,t)=E_{SS}+E_{SI}+E_{IS}+E_{II}.
\end{split}
\end{equation}
The overall energy function is determined by the two kinds of entity representations jointly. $E_{SS}=||\mathbf{h}_S+\mathbf{r}-\mathbf{t}_S||$ is the same energy function as TransE which only depends on the structure-based representations. $E_{II}=||\mathbf{h}_I+\mathbf{r}-\mathbf{t}_I||$ is the energy function in which both head and tail entities are image-based representations learned from their corresponding images. We also have $E_{SI}=||\mathbf{h}_S+\mathbf{r}-\mathbf{t}_I||$ and $E_{IS}=||\mathbf{h}_I+\mathbf{r}-\mathbf{t}_S||$ to assure that both structure-based representations and image-based representations are learned into the same vector space.

\begin{figure*}[!htbp]
\centering
\includegraphics[width=0.84\textwidth]{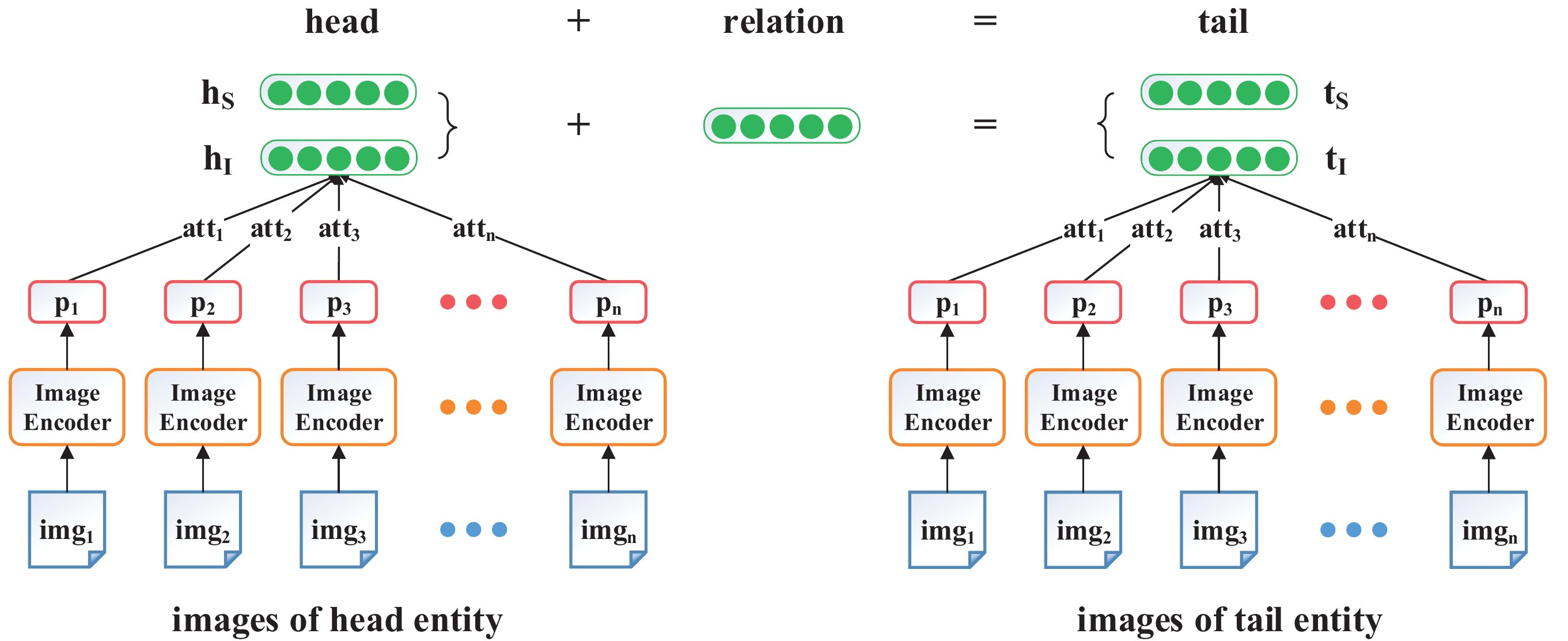}
\caption{Overall architecture of the IKRL model.}\label{fig. 2}
\end{figure*}

According to the energy function, the overall architecture of the IKRL model is demonstrated in Fig. \ref{fig. 2}. Each entity has multiple entity images providing significant visual information. First, we design a neural image encoder taking every entity image as inputs. The image encoder aims to extract informative features from images and construct the image representations in entity space. Second, in order to combine multiple image representations, we implement an instance-level attention-based learning method to automatically calculate the attention we should pay on different image instances for each entity. Finally, the aggregated image-based representations are learned jointly combined with the structure-based representations under the overall energy function.

\begin{figure}[!htbp]
\centering
\includegraphics[width=0.95\columnwidth]{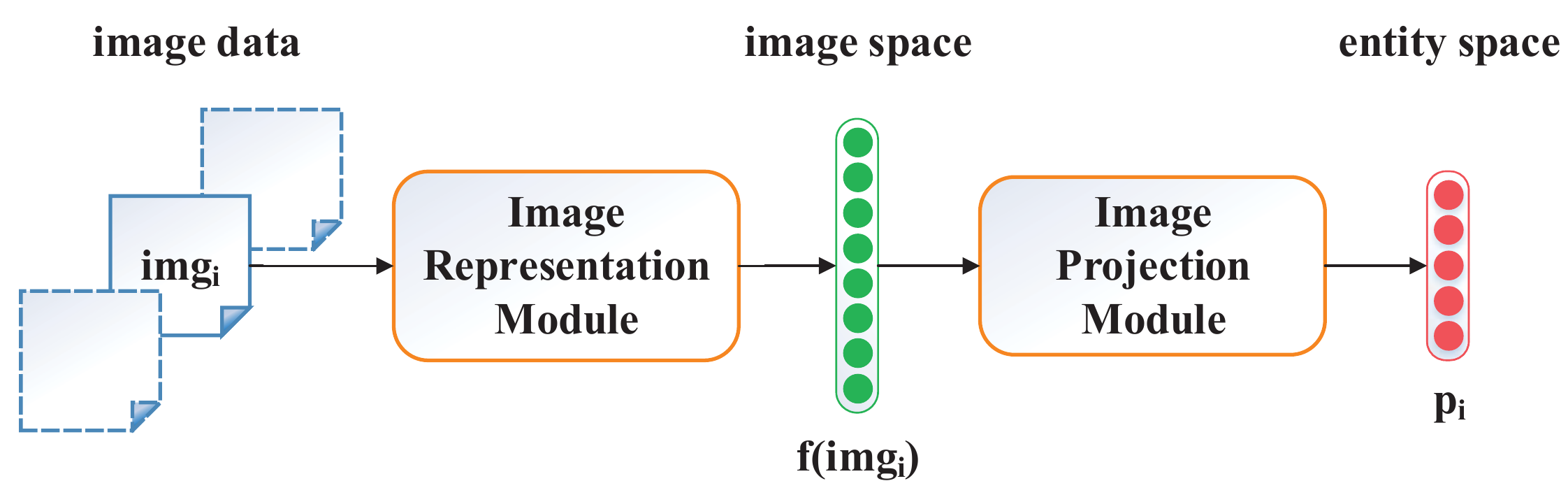}
\caption{Image encoder.}\label{fig. 3}
\end{figure}

\subsection{Image Encoder}

Images provide informative visual information that can intuitively describe the appearances and behaviours of entities, which are considered as the fundamental input data of the IKRL model. For each entity $e_k$, there are multiple image instances represented as $I_k=\{img_1^{(k)}, img_2^{(k)}, \cdots , img_n^{(k)}\}$.

To effectively encode the image information into knowledge representations, we propose an image encoder which consists of an image representation module and an image projection module. The image representation module utilizes neural networks to extract discriminative features in images, and constructs image feature representations for each image. Next, the image projection module attempts to project those image feature representations from image space to entity space. Fig. \ref{fig. 3} demonstrates the overall pipeline of the image encoder.

\subsubsection{Image Representation Module}

The image representation module aims to build the image feature representations. We utilize AlexNet, a widely-used neural network that contains five convolution layers, two fully-connected layers and a softmax layer, to extract image features \cite{krizhevsky2012imagenet}. During preprocessing, all images are reshaped to $224 \times 224$ from the center, corners and their horizontal reflections. Inspired by \cite{shutovablack2016black}, we take the 4096-dimensional embeddings which are outputs of the second fully-connected layer (also called as fc7) as the image feature representations.

\subsubsection{Image Projection Module}

After we get the compressed feature representations for each image, the next procedure is to build the bridges between images and entities via image projection module. Specifically, we transfer the image feature representations from image space to entity space with a shared projection matrix. The image-based representation $\mathbf{p}_{i}$ in entity space for the $i$-th image is defined as:
\begin{equation}
\begin{split}
\mathbf{p}_{i}=\mathbf{M} \cdot f(img_i),
\end{split}
\end{equation}
in which $\mathbf{M} \in \mathbb{R}^{d_i \times d_s}$ is the projection matrix, $d_i$ represents the dimension of image features, while $d_s$ represents the dimension of entities. $f(img_i)$ stands for the $i$-th image feature representation in image space, which is constructed by the image representation module.

\subsection{Attention-based Multi-instance Leaning}

Image encoder takes images as inputs and then constructs image-based representations for each single image. However, most entities have more than one image from different aspects in various scenarios. It is essential but also challenging to determine which images are better to represent their corresponding entities. Simply summing up all image representations may suffer from noises and lose detailed information. Instead, to construct the aggregated image-based representation for each entity from multiple instances, we propose an attention-based multi-instance learning method.

Attention-based methods confirm to be intelligent to automatically select informative instances form multiple candidates. It has been widely utilized in various fields such as image classification \cite{mnih2014recurrent}, machine translation \cite{bahdanau2015neural} and abstractive sentence summarization \cite{rush2015neural}. We jointly consider each image representation and the structure-based representation of its corresponding entity to generate the instance-level attention. For the $i$-th image representation $\mathbf{p}_{i}^{(k)}$ of the $k$-th entity, the attention is defined as follows:
\begin{equation}
\begin{split}
att(\mathbf{p}_{i}^{(k)}, \mathbf{e}_S^{(k)})=\frac{\exp{(\mathbf{p}_{i}^{(k)} \cdot \mathbf{e}_S^{(k)})}}{\sum_{j=1}^{n}{\exp{(\mathbf{p}_{j}^{(k)} \cdot \mathbf{e}_S^{(k)})}}},
\end{split}
\end{equation}
where $\mathbf{e}_S^{(k)}$ represents the structure-based representation of the $k$-th entity. High attention indicates that the image representation is similar to its corresponding structure-based representation, and thus should be more considered when building the aggregated image-based representation according to the energy function. Hence, we define the aggregated image-based representation for the $k$-th entity as follows:
\begin{equation}
\begin{split}
\mathbf{e}_I^{(k)}=\sum_{i=1}^{n}\frac{att(\mathbf{p}_{i}^{(k)}, \mathbf{e}_S^{(k)}) \cdot \mathbf{p}_{i}^{(k)}}
{\sum_{j=1}^{n}att(\mathbf{p}_{j}^{(k)}, \mathbf{e}_S^{(k)})},
\end{split}
\end{equation}

Besides the attention-based method, we also implement two alternative combination methods for further comparisons. AVG is a simple combination method that takes the mean of all image embeddings, supposing that every image has equal contributions to the final image-based representation. MAX is a simplified version for attention, which only considers the image representations with the highest attention.

\subsection{Objective Formalization}

We utilize a margin-based score function as our training objective, which is defined as follows:
\begin{equation}
\begin{split}
L=\sum_{(h,r,t)\in T}\sum_{(h',r',t')\in T'}\max(\gamma+E(h,r,t)-\\
E(h',r',t'),0),
\end{split}
\end{equation}
where $\gamma$ is a margin hyperparameter. $E(h,r,t)$ is the overall energy function stated above, in which both head and tail entities have two kinds of representations including structure-based representations and image-based representations. $T'$ stands for the negative sample set of $T$, which we define as follows:
\begin{equation}
\begin{split}
T'=\{(h',r,t)|h'\in E\}\cup\{(h,r,t')|t'\in E\}\cup\\
\{(h,r',t)|r'\in R\}, \quad (h,r,t)\in T,
\end{split}
\end{equation}
which means one of the entities or relation in a triple has been randomly replaced by another one. We also wipe out all generated negative triples that are already in $T$ to assure triples in $T'$ are truly negative.

\subsection{Optimization and Implementation Details}

The IKRL model can be formalized as a parameter set $\theta=(\mathbf{E}, \mathbf{R}, \mathbf{W}, \mathbf{M})$, in which $\mathbf{E}$ stands for the structure-based embedding set of entities, $\mathbf{R}$ stands for the embedding set of relations. $\mathbf{W}$ represents the weights of the neural networks used in image representation module, while $\mathbf{M}$ represents the projection matrix used in image projection module.

We utilize mini-batch stochastic gradient descent (SGD) to optimize our model, with chain rule applied to update the parameters. $\mathbf{M}$ is initialized randomly, while $\mathbf{E}$ and $\mathbf{R}$ could be either initialized randomly or be pre-trained by previous translation-based methods. As for the image representation module, we utilize a deep learning framework Caffe \cite{jia2014caffe} for image representation, which is pre-trained on ILSVRC 2012 with a minor variation from the version described in \cite{krizhevsky2012imagenet}. The weights of AlexNet $\mathbf{W}$ are pre-trained and fixed during training. For the consideration of efficiency, we use GPU to accelerate image representation, and also employ a multi-thread version for training.

\section{Experiments}

\subsection{Dataset}

In this paper, we construct a new dataset of knowledge graph combined with images named WN9-IMG for evaluation tasks including knowledge graph completion and triple classification. The triple part of WN9-IMG is the subset of a classical KG dataset WN18 \cite{bordes2014semantic}, which is originally extracted from WordNet \cite{miller1995wordnet}. For the consideration of image quality, we use 63,225 images extracted from ImageNet \cite{deng2009imagenet}, which is a huge image database organized according to the WordNet hierarchy. We assure that all entities in WN9-IMG should have images, and randomly split extracted triples into train, validation and test set. The statistics of WN9-IMG are listed in Table \ref{tab. 1}.

\begin{table}[!htbp]
\center
\small
\caption{\label{tab. 1} Statistics of the dataset}
\begin{tabular}{cccccc}
 \toprule
  Dataset & \#Rel & \#Ent & \#Train & \#Valid & \#Test\\
 \midrule
  WN9-IMG & 9 & 6,555 & 11,741 & 1,337 & 1,319\\
 \bottomrule
\end{tabular}

\end{table}

\subsection{Experiment Settings}

We train the IKRL model via mini-batch SGD, with the margin $\gamma$ set among $\{1.0, 2.0, 4.0\}$. The learning rate $\lambda$ could be either empirically fixed among $\{0.0002, 0.0005, 0.001\}$, or designed following a flexible adaptive strategy that descends through iterations. The optimal configurations of the IKRL model are: $\gamma=4.0$, with the learning rate defined adopting a linear-declined strategy in which $\lambda$ ranges form $0.001$ to $0.0002$. To balance efficiency and diversity, the image number $n$ for all entity is up to $10$. We also set the dimension of image feature embeddings $d_i=4096$, and the dimension of entity and relation embeddings $d_s=50$.

We implement TransE \cite{bordes2013translating} and TransR \cite{lin2015learning} as our baselines following the same experimental settings reported in their papers. For fair comparisons, the dimensions of entities and relations in all baselines are also set to be $50$.

\begin{table*}[!htbp]
\center
\small
\caption{\label{tab. 3} Evaluation results on different combination strategies}
\begin{tabular}{p{2.5cm}<{\centering}|p{0.8cm}<{\centering}p{0.8cm}<{\centering}|p{0.8cm}<{\centering}p{0.8cm}<{\centering}|p{0.8cm}<{\centering}p{0.8cm}<{\centering}|p{0.8cm}<{\centering}p{0.8cm}<{\centering}}
 \toprule
  Type & \multicolumn{4}{c|}{Image-based representation} & \multicolumn{4}{c}{Structure-based representation} \\
  \midrule
  \multirow{2}{*}{Metric}
   & \multicolumn{2}{c|}{Mean Rank} & \multicolumn{2}{c|}{Hits@10(\%)} & \multicolumn{2}{c|}{Mean Rank} & \multicolumn{2}{c}{Hits@10(\%)} \\
   & Raw & Filter & Raw & Filter & Raw & Filter & Raw & Filter \\
  \midrule
  IKRL (MAX) & 59 & 52 & 79.8 & 92.1 & 62 & 55 & 81.0 & 92.3\\
  IKRL (AVG) & \textbf{29} & \textbf{22} & 79.3 & 92.9 & 43 & 36 & 80.7 & 92.8\\
  IKRL (ATT) & \textbf{29} & \textbf{22} & \textbf{80.2} & \textbf{93.3} & \textbf{41} & \textbf{34} & \textbf{81.1} & \textbf{92.9}\\
 \bottomrule
\end{tabular}
\end{table*}

\subsection{Knowledge Graph Completion}

\subsubsection{Evaluation Protocol}

The task of knowledge graph completion aims to complete a triple $(h, r, t)$ when one of $h$, $r$, $t$ is missing. We mainly focus on entity prediction in evaluation. This task has been widely used to evaluate the quality of knowledge representations \cite{bordes2012joint,bordes2013translating}. The prediction is determined via the dissimilarity function $||\mathbf{h}+\mathbf{r}-\mathbf{t}||$. Since the IKRL model has two kinds of representations, we will report three prediction results based on our models: IKRL (SBR) only utilizes structure-based representations for all entities when predicting the missing ones, while IKRL (IBR) only utilizes image-based representations for prediction. IKRL (UNION) is a simple joint method considering the weighted concatenation of both entity representations.

Following the same settings in \cite{bordes2013translating}, we consider two measures as our evaluation metrics in entity prediction: (1) mean rank of correct entities (Mean Rank); (2) proportion of correct entity results ranked in top 10 (Hits@10). We also follow the two evaluation settings named ``Raw" and ``Filter" used in \cite{bordes2013translating}. In this section, we first demonstrate the results of entity prediction, and then implement another experiment for further discussions on the power of attention.

\begin{table}[!htbp]
\center
\small
\caption{\label{tab. 2} Evaluation results on entity prediction}
\begin{tabular}{c|cc|cc}
 \toprule
  \multirow{2}{*}{Metric} & \multicolumn{2}{c|}{Mean Rank} & \multicolumn{2}{c}{Hits@10(\%)} \\
   & Raw & Filter & Raw & Filter\\
 \midrule
  TransE & 143 & 137 & 79.9 & 91.2\\
  TransR & 147 & 140 & 80.1 & 91.7\\
  \midrule
  IKRL (SBR) & 41 & 34 & \textbf{81.1} & 92.9\\
  IKRL (IBR) & 29 & 22 & 80.2 & 93.3\\
  IKRL (UNION) & \textbf{28} & \textbf{21} & 80.9 & \textbf{93.8}\\
 \bottomrule
\end{tabular}
\end{table}

\subsubsection{Entity Prediction}

The results of entity prediction are demonstrated in Table \ref{tab. 2}. From the results we can observe that: (1) all IKRL models outperform all baselines on both evaluation metrics of Mean Rank and Hits@10, among which IKRL (UNION) achieves the best performance. It indicates that the visual information of images has been successfully encoded into entity representations, which is of great significance when constructing knowledge representations. (2) Both IKRL (SBR) and IKRL (IBR) have better performances compared to the baselines, which indicates that visual information could not only instruct the construction of image-based representations, but also improve the performances of structure-based representations. (3) The IKRL models significantly and consistently outperform baselines on Mean Rank. It is because that Mean Rank depends on the overall quality of knowledge representations, and thus is sensitive to the wrong-predicted results. Previous translation-based methods like TransE only consider the structured information in triples, which may fail to predict the relationships if the corresponding information is missing. However, the image information utilized in IKRL can provide supplementary information. Therefore, the results of IKRL are much better than baselines on Mean Rank.

\subsubsection{Further Discussion on Attention}

To further demonstrate the power of attention-based methods, we implement three combination strategies to jointly consider multiple image instances. IKRL (ATT) represents the basic model with attention when constructing the aggregated image-based representations, while IKRL (MAX) represents the combination strategy that only considers the image instance which has the largest attention, and IKRL (AVG) represents the strategy that takes the average embeddings of all image instances to represent an entity. The evaluation results on entity prediction with both image-based representations and structure-based representations are shown in Table \ref{tab. 3}.

From Table \ref{tab. 3} we observe that: (1) all IKRL models still outperform baselines on Mean Rank and Hits@10 no matter what the combination strategy is. It confirms the improvements introduced by images, for the visual information has been successfully encoded into knowledge representations. (2) The IKRL (ATT) model achieves the best performance among all three combination strategies, which implies that the attention-based method is capable of automatically selecting more informative image instances to represent entities. (3) The IKRL (AVG) model performs better than the IKRL (MAX) model, which indicates that only considering images with the largest attention will lose important information located in other instances. (4) It seems that the IKRL (ATT) model only has slight advantages over the IKRL (AVG) model. The reason is that the qualities of images we extract from ImageNet are very high, which may narrow the gap between attention-based and average-based methods. For further analysis, we will give some examples with attention in case study, which could successfully distinguish the relatively better and worse images from all candidates.

\subsection{Triple Classification}

\subsubsection{Evaluation Protocol}

Triple classification aims to predict whether a triple fact $(h, r, t)$ is correct or not according to the dissimilarity function \cite{socher2013reasoning}. It can be viewed as a binary classification task on triples. Since WN9-IMG has no explicit negative instances, we generate the negative instances by randomly replacing head or tail entities with another entity following the same protocol utilized in \cite{socher2013reasoning}. We also assure that the number of positive triples is equal to that of negative triples.

In classification, we set different relation-specific thresholds $\delta_r$ for each relation, which are optimized by maximizing the classification accuracies on the validation set with their corresponding relations. For a triple to be classified, if its dissimilarity function $||\mathbf{h}+\mathbf{r}-\mathbf{t}||$ is over $\delta_r$, it will be predicted to be negative, and otherwise to be positive. To better demonstrate the advantages in IKRL models, only image-based representations are utilized when calculating the dissimilarity function for each triple.

\begin{table}[!htbp]
\center
\small
\caption{\label{tab. 4} Evaluation results on triple classification}
\begin{tabular}{p{3cm}<{\centering}p{2cm}<{\centering}}
 \toprule
  Methods & Accuracy(\%)\\
 \midrule
  TransE & 95.0\\
  TransR & 95.3\\
 \midrule
  IKRL (MAX) & 96.3\\
  IKRL (AVG) & 96.6\\
  IKRL (ATT) & \textbf{96.9}\\
 \bottomrule
\end{tabular}
\end{table}

\begin{figure*}[!htbp]
\centering
\includegraphics[width=0.84\textwidth]{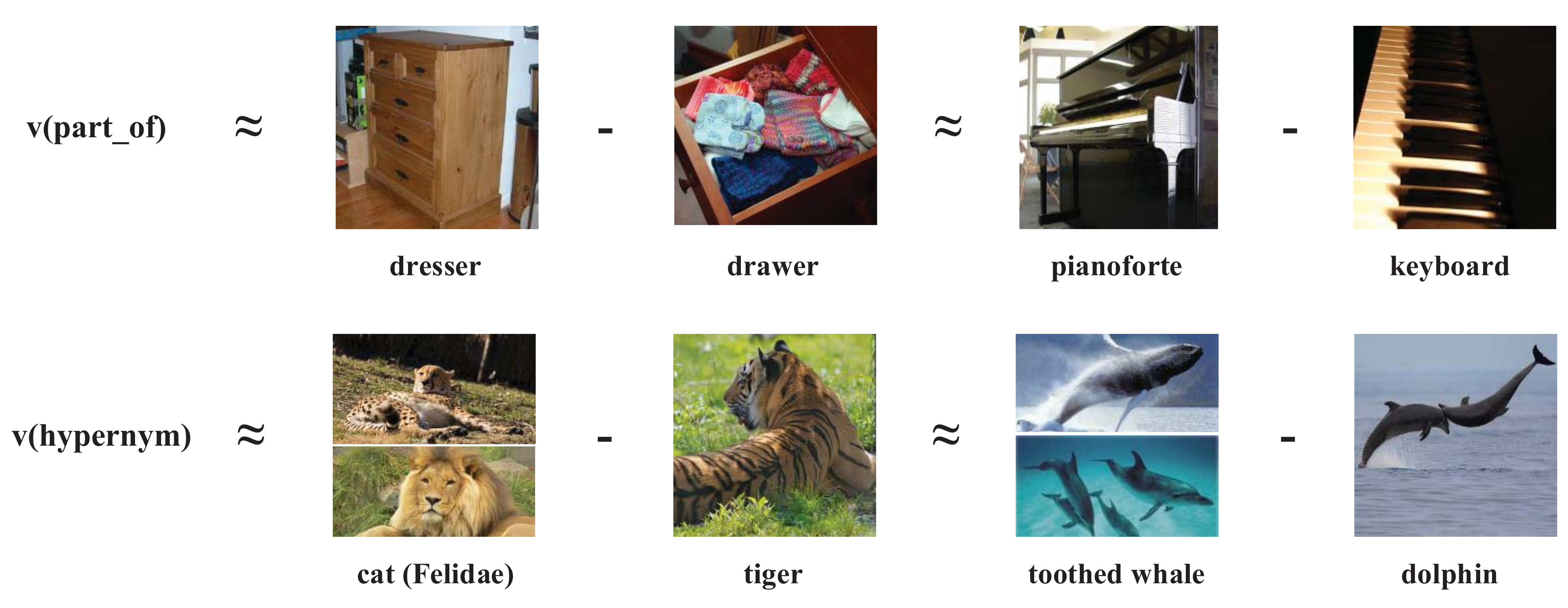}
\caption{Examples of semantic regularities on images.}\label{fig. 4}
\end{figure*}

\subsubsection{Experimental Results}

From Table \ref{tab. 4} we can observe that: (1) all IKRL models outperform both baselines, which demonstrates the effectiveness and robustness of our models that combine structured information in triples with visual information in images. Note that IKRL is based on the framework of TransE but still performs better even when compared with the enhanced TransR model, which confirms the improvements introduced by images. (2) The IKRL (ATT) model achieves the best performance compared to other combination strategies. It indicates that the attention-based method can jointly take multiple instances into consideration and smartly choose more informative images from all candidates.

\subsection{Case study}

In this section, we give two cases with detailed analysis. The first is to introduce the semantic regularities of images, and the second is to demonstrate the capability of attention. For better demonstrations, the images shown in case study may be chopped while the main objects are included.

\subsubsection{Semantic Regularities of Images}

\cite{mikolov2013linguistic} shows that word representations have some interesting semantic regularities such as $v(king)-v(man) \approx v(queen)-v(woman)$, in which $v(x)$ indicates the word embedding of $x$. These similar regularities have also been found in image-text space \cite{kiros2014unifying}. In image-knowledge joint space, we explore such semantic translation regularities on image-based representations demonstrated in Fig. \ref{fig. 4}. Differing from previous work, the result of \emph{dresser} minus \emph{drawer} matches a concrete and meaningful relation \texttt{part\_of}, which makes the semantic translation regularities in images-knowledge space more interpretable.

\subsubsection{Capability of Attention}

\begin{figure}[!htbp]
\centering
\includegraphics[width=0.99\columnwidth]{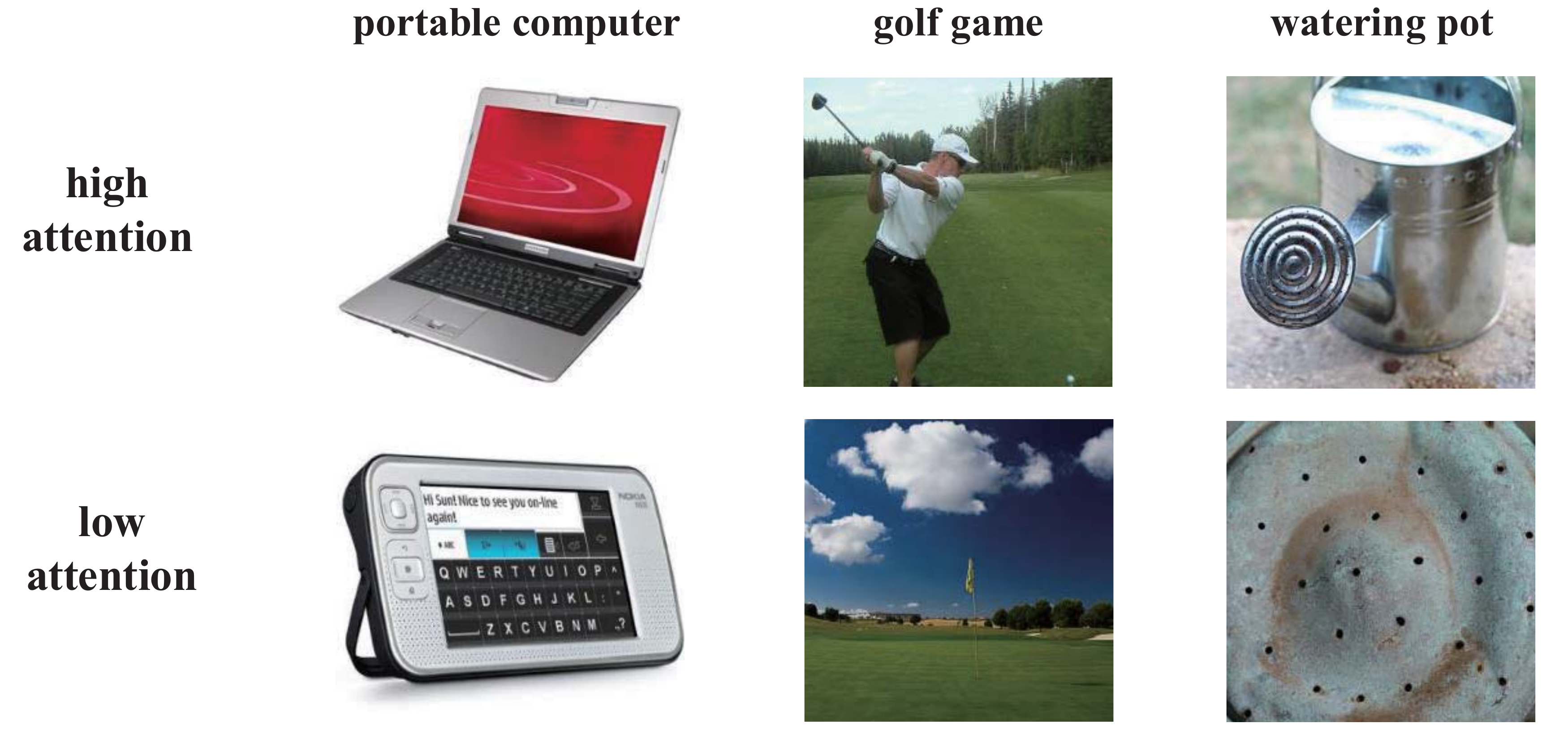}
\caption{Examples of images with different attention.}\label{fig. 5}
\end{figure}

Fig. \ref{fig. 5} demonstrates some pairs of image instances with different attention, aiming to confirm the capability of attention in selecting more informative images from multiple instances. In the first example of \emph{portable computer}, the attention-based method successfully detects the low-quality instance which is actually a phone by assigning low attention. For \emph{golf game}, the image with low attention only shows an overview of the lawn without any person or detailed sporting goods, and thus is less considered in combination. As for \emph{watering pot}, the low-attention image concentrates on the spout of a watering pot, which will be confusing to represent the whole entity. With the help of attention, we can automatically learn knowledge representations from better images, alleviating the noises in multiple image instances.

\section{Conclusion and Future Work}

In this paper, we propose the IKRL models learning knowledge representations with images. We utilize neural networks and a projection module to model each image, and then construct the aggregated image-based representations by combining multiple image instances based on attention. Experimental results confirm that our models are capable of encoding image information into knowledge representations. The source code and dataset of this paper can be obtained from \url{https://github.com/thunlp/IKRL}.

We will explore the following research directions in future: (1) the quality of image representations is essential. We will utilize more sophisticated models to extract better image features in some specific domains. (2) The current IKRL models are based on TransE. We will explore the effectiveness of our models when extended to other enhanced translation-based methods. (3) In this paper, we only consider each entity image as the visual representation of its target entity, while sometimes an image contains far beyond a simple entity. We will explore to learn multiple entities and their relations within one image combined with our IKRL models.

\section*{Acknowledgments}

This work is supported by the Natural Science Foundation of China (NSFC) and the German Research Foundation (DFG) in Project Crossmodal Learning, NSFC 61621136008 / DFC TRR-169, and the National Natural Science Foundation of China (NSFC No. 61661146007, 61572273), and Tsinghua University Initiative Scientific Research Program (20151080406).

\bibliographystyle{named}
\bibliography{reference}

\end{document}